\documentclass{article}
\usepackage{spconf,amsmath,amssymb,graphicx}
\usepackage[hidelinks]{hyperref}
\usepackage{tikz,booktabs}
\usepackage{pgfplots}
\pgfplotsset{compat=1.18}
\usepgfplotslibrary{groupplots}

\title{Personalised federated learning for Riemannian and Euclidean EEG decoding}

\name{Thibault Pautrel$^{1}$ \qquad Florent Bouchard$^{1}$ \qquad
      Ammar Mian$^{2}$ \qquad Guillaume Ginolhac$^{2}$}
\address{$^{1}$Universit\'{e} Paris-Saclay, CNRS, CentraleSup\'{e}lec, L2S, France \\
         $^{2}$Universit\'{e} Savoie Mont-Blanc, LISTIC, France}
         
\usetikzlibrary{arrows.meta,calc,positioning,fit,decorations.pathreplacing,
                backgrounds}

\newsavebox{\archbox}

\tikzset{
  netbox/.style={draw=black,semithick,rounded corners=2pt,align=center,
                 fill=white,font=\scriptsize,minimum height=7mm,inner sep=2pt},
  netdata/.style={netbox,dashed},
  netarrow/.style={-{Latex[length=1.6mm]},semithick,draw=black},
  biarrow/.style={{Latex[length=1.6mm]}-{Latex[length=1.6mm]},semithick,
                  draw=black},
  shp/.style={midway,above,font=\tiny,inner sep=1pt},
  rowname/.style={font=\scriptsize\bfseries,anchor=south west},
  agg/.style={netbox,fill=black!10},
  frz/.style={netbox,fill=black!28,densely dotted},
  grp/.style={draw=black,thin,rounded corners=3pt,inner sep=2pt},
}

\newcommand{\ArchFigure}{%
\begin{figure*}[t]
\centering
\sbox{\archbox}{%
\begin{tikzpicture}[x=1mm,y=1mm,
    trunkbox/.style={netbox,fill=black!15}]
  \def\rowtwo{-15}                       

  \node[netdata,minimum width=13mm] (scm)  at (  6.5,0) {SCM $\Sigma$};
  \node[trunkbox,minimum width=16mm] (bim)  at ( 35  ,0) {BiMap\\$W^{\top}\!\Sigma W$};
  \node[trunkbox,minimum width=13mm] (re)   at ( 61.5,0) {ReEig\\$\varepsilon$};
  \node[trunkbox,minimum width=13mm] (log)  at ( 93.5,0) {LogEig};
  \node[trunkbox,minimum width=13mm] (vech) at (118.5,0) {Vech};
  \node[netbox ,minimum width=11mm] (fc)   at (149.5,0) {FC};
  \node[netdata,minimum width=11mm] (yh)   at (172.5,0) {$\hat{y}$};

  \draw[netarrow] (scm)  -- node[shp]{$(B,C,C)$}         (bim);
  \draw[netarrow] (bim)  -- node[shp]{$(B,d,d)$}         (re);
  \draw[netarrow] (re)   -- node[shp]{$(B,d,d)$}         (log);
  \draw[netarrow] (log)  -- node[shp]{$(B,d,d)$}         (vech);
  \draw[netarrow] (vech) -- node[shp]{$(B,d(d{+}1)/2)$}  (fc);
  \draw[netarrow] (fc)   -- node[shp]{$(B,K)$}           (yh);

  \node[rowname] at (0,4.2) {SPDNet};

  \node[netdata,minimum width=13mm] (tri) at (  6.5,\rowtwo) {trial $X$};
  \node[trunkbox,minimum width=41mm] (b1)  at ( 47.5,\rowtwo) {temporal \& spatial convolution\\\tiny batch norm, ELU, pooling, dropout};
  \node[trunkbox,minimum width=38mm] (b2)  at (106  ,\rowtwo) {separable convolution\\\tiny batch norm, ELU, pooling, dropout};
  \node[netbox ,minimum width=11mm] (efc) at (149.5,\rowtwo) {FC};
  \node[netdata,minimum width=11mm] (eyh) at (172.5,\rowtwo) {$\hat{y}$};

  \draw[netarrow] (tri) -- node[shp]{$(B,1,C,\mathcal{T})$}      (b1);
  \draw[netarrow] (b1)  -- node[shp]{$(B,F_1D,1,\mathcal{T}/4)$} (b2);
  \draw[netarrow] (b2)  -- node[shp]{$(B,F_2,1,\mathcal{T}/32)$} (efc);
  \draw[netarrow] (efc) -- node[shp]{$(B,K)$}                    (eyh);

  \node[rowname] at (0,\rowtwo+5.5) {EEGNet};

  \begin{scope}[on background layer]
    \fill[black!10,rounded corners=3pt] (25,4.3) rectangle (127,-4.3);
    \fill[black!10,rounded corners=3pt] (25,\rowtwo+5.6) rectangle (127,\rowtwo-5.6);
    \node[grp,fit=(bim)(re)] (bire) {};
  \end{scope}

  \node[font=\tiny,anchor=south] at (48.2,-7.6) {BiRe block, stacked $L$ times};
  \node[font=\scriptsize,anchor=south] at (75,5.0) {trunk};
  \node[font=\scriptsize,anchor=south] at (149.5,5.0) {head};
\end{tikzpicture}}%
\ifdim\wd\archbox>\linewidth
  \resizebox{\linewidth}{!}{\usebox{\archbox}}%
\else
  \usebox{\archbox}%
\fi
\caption{The two architectures, each split into a trunk (shaded) and a linear
head. SPDNet maps the covariance matrix through $L$ BiRe blocks, then to a
linear space. EEGNet applies $F_1=8$ temporal filters, a depthwise spatial
convolution with $D=2$ filters per temporal filter and a separable convolution giving $F_2=16$ feature maps, with
average pooling by $4$ and then by $8$. Tensor shapes are given above the
arrows, with $B$ the batch size and $d$ the BiMap output dimension.}
\label{fig:arch}
\end{figure*}}

\begin{document}
\maketitle
\begin{abstract}
Federated learning (FL) lets EEG decoders learn from recordings of several
subjects without pooling them. We consider two light EEG
decoders, the Riemannian SPDNet and the Euclidean EEGNet. Both split into a
trunk, which builds a latent representation, and a head, which classifies
it. Inter-subject variability, however, makes a
single shared FL model a poor fit for each subject. Personalised FL
addresses this: all subjects learn a common trunk, and each subject keeps
its own head. We adapt it for SPDNet and study its effects against standard FL and centralised
training, with EEGNet as a Euclidean baseline. Experiments cover three
motor-imagery datasets that span diverse regimes in channels, subjects and
classes. We observe that personalised SPDNet reaches higher
accuracy than both standard FL and centralised training, while converging
in fewer rounds and communicating fewer parameters than standard FL. It
also outperforms every EEGNet configuration on two of the three datasets,
although centralised EEGNet outperforms centralised SPDNet.
\end{abstract}

\section{Introduction}

Electroencephalography (EEG) records brain activity through electrodes
placed on the scalp. An EEG decoder maps a multi-channel trial to a class,
such as a mental command in a brain-computer interface
(BCI)~\cite{congedo2017primer}, a cognitive state or a clinical event.
Two families of decoders are common. Euclidean architectures such as EEGNet~\cite{lawhern2018eegnet}
apply convolutions to the raw trials. Riemannian architectures such as
SPDNet~\cite{huang2017spdnet} and its
variants~\cite{wilson2025deep,carrara2025geometric} process the spatial
covariance matrix of each trial, which is symmetric positive definite (SPD),
while preserving its geometry.

EEG recordings are collected subject by subject, often at different sites.
They contain medical information that regulation and confidentiality keep
from being pooled. Federated learning (FL)~\cite{mcmahan2017communication}
fits this setting. Several clients train a single model under the
coordination of a central server and exchange parameters instead of raw
trials. At each round, every client trains the model on its own data and the
server aggregates the returned parameters. The decoder thus learns from all
recordings, and no recording leaves its site. FL has already been applied
to EEG decoding~\cite{liu2024aggregating}, including through
transfer learning on covariance matrices~\cite{jia2024federated}.
Euclidean architectures fit directly into standard FL. Clients run
stochastic gradient descent and the server averages the returned weights
(FedAvg). A Riemannian architecture needs both steps adapted to the
geometry. Clients need manifold-aware optimisation, and the server needs an
aggregation rule whose output stays on the manifold. A first federated
framework for SPDNet combined Riemannian optimisation on the clients with
geometry-aware aggregation on the server~\cite{pautrel2026fedspdnet}, such
as ProjAvg, which projects the average of the client weights back onto the manifold. It showed that SPDNet can be federated, with limited gains. Indeed, federated SPDNet improved little over federated EEGNet, and both stayed below their centralised
counterparts trained on pooled data.

A likely cause of this gap is the high inter-subject variability of
EEG~\cite{saha2020intra}. When each client is one subject, the clients may face
a domain shift, and a single shared model may fit none of them well.
In the Riemannian setting, this shift appears directly in
the covariance matrices, and subject-specific recentering of these matrices
is an effective way to transfer decoders across
subjects~\cite{zanini2017transfer}. Personalised FL addresses such
variability by keeping part of the model on each
client~\cite{tan2022towards}. Both SPDNet and EEGNet split
into a trunk that builds a representation and a linear head that classifies
it. FedPer~\cite{arivazhagan2019fedper} and FedRep~\cite{collins2021fedrep}
exploit this split. They share the trunk across clients and keep the head
local. These methods were designed
for Euclidean networks. Whether they
transfer to a Riemannian network like SPDNet, and whether both families
benefit in the same way, remains open.

\noindent\textbf{Contributions.} (i) We adapt and implement personalised FL
with a local head for SPDNet, and aggregate its trunk with ProjAvg. (ii) We evaluate it
on three motor-imagery datasets, alongside the Euclidean baseline EEGNet,
against standard FL and centralised training. (iii) We study the effects of
personalisation on accuracy, convergence and communication.
We observe that, for SPDNet, personalised FL outperforms both standard FL
and centralised training on all three datasets. It also converges in fewer
rounds and transmits fewer parameters than standard FL. For EEGNet,
personalised FL improves on standard FL on two of the three datasets, but
stays below centralised training and brings no speed-up.

Section~\ref{sec:method} presents the architectures and the federated
protocols. Section~\ref{sec:design} describes the experimental design.
Section~\ref{sec:analysis} analyses the results and
Section~\ref{sec:conclusion} concludes.
\section{Method}\label{sec:method}
\subsection{EEG decoding architectures}
An EEG trial is a labelled record $(X,y)$, where $X\in\mathbb{R}^{C\times\mathcal{T}}$
holds $\mathcal{T}$ time samples on $C$ channels and $y\in\{1,\dots,K\}$ is the
corresponding class, typically encoding movements like left hand, right hand, tongue or feet. 

EEGNet~\cite{lawhern2018eegnet} is a compact convolutional network applied to the raw trials. A first block combines temporal convolutions with a depthwise spatial convolution that learns several spatial filters per temporal feature. A second block applies a separable convolution, factorised into a depthwise temporal kernel followed by pointwise filters. Both blocks use batch normalisation, ELU activation, average pooling and dropout, and a linear layer produces the class scores.
\ArchFigure

Alternatively to raw signals, popular features for BCI pipelines are covariance matrices as input features~\cite{barachant2012multiclass,congedo2017primer}. For a centered trial $\bar{X}$, the spatial covariance matrix $\Sigma=\frac{1}{\mathcal{T}-1}\bar{X}\bar{X}^{\top}$ is symmetric positive definite (SPD).
SPDNet~\cite{huang2017spdnet} maps a covariance matrix to class scores
while preserving positive definiteness. It stacks $L$ \textbf{BiRe} blocks.
Block $\ell$ takes a $d_{\ell-1}\times d_{\ell-1}$ SPD matrix
$\Sigma_{\ell-1}$, with $\Sigma_0=\Sigma$ and $d_0=C$, and returns a
$d_\ell\times d_\ell$ SPD matrix $\Sigma_\ell$, with $d_\ell\le d_{\ell-1}$.
It chains two layers. The \textbf{BiMap} layer
$\Sigma_{\ell-1}\mapsto W_\ell^{\top}\Sigma_{\ell-1}W_\ell$ reduces the
dimension. Its weight lies on the Stiefel manifold
$\mathrm{St}_{d_{\ell-1},d_{\ell}}=\{W\in\mathbb{R}^{d_{\ell-1}\times d_{\ell}}:\,W^{\top}W=I_{d_{\ell}}\}$,
so the output stays SPD. The \textbf{ReEig} layer adds a nonlinearity. For
$S=U\Lambda U^{\top}$, $\mathrm{ReEig}_{\varepsilon}(S)=U\max(\varepsilon I,\Lambda)U^{\top}$
raises every eigenvalue below the threshold $\varepsilon>0$ to $\varepsilon$.
The block output is thus
$\Sigma_\ell=\mathrm{ReEig}_{\varepsilon}(W_\ell^{\top}\Sigma_{\ell-1}W_\ell)$.
A \textbf{LogEig} layer then maps $\Sigma_L$ to the space of symmetric
matrices by taking the logarithm of its eigenvalues. A softmax classifier
reads the half-vectorisation $\mathrm{vech}(\log\Sigma_L)$, which stacks
its $d_L(d_L+1)/2$ upper-triangular entries, giving
$\hat{y}=\mathrm{softmax}(\xi\,\mathrm{vech}(\log\Sigma_L)+\beta)$ with
$\xi\in\mathbb{R}^{K\times d_L(d_L+1)/2}$ and $\beta\in\mathbb{R}^{K}$.

Both architectures therefore split as
$\theta=(\theta_{\mathrm{trunk}},\theta_{\mathrm{head}})
\in\mathcal{M}_{\mathrm{trunk}}\times\mathcal{M}_{\mathrm{head}}$,
with a trunk that builds the representation and a head that classifies it. The head is the softmax layer $(\xi,\beta)$ for SPDNet and the final linear layer for EEGNet, Euclidean in both cases. The trunks differ: that of SPDNet, $\theta_{\mathrm{trunk}}=(W_1,\dots,W_L)$, lies on a product of Stiefel manifolds, while that of EEGNet gathers the convolutional and batch-normalisation parameters and lies in a Euclidean space.

\subsection{Federated protocols}
\label{ssec:fl}
\indent Each client $i=1,\dots,N$ holds a dataset $\mathcal{D}^{(i)}$ of labelled
samples $z=(x,y)$, where the input $x$ is the raw trial $X$ for EEGNet and the
covariance matrix $\Sigma$ for SPDNet. Its empirical risk is
$F_i(\theta)=|\mathcal{D}^{(i)}|^{-1}\sum_{z\in\mathcal{D}^{(i)}}\ell(\theta;z)$,
with the cross-entropy loss $\ell(\theta;z)=-\log[f_{\theta}(x)]_{y}$ and
$f_{\theta}$ the forward pass of the architecture.

\textbf{Standard FL.} Standard federated learning seeks a single model minimising
$F(\theta)=\frac{1}{N}\sum_{i=1}^{N}F_i(\theta)$ over
$\mathcal{M}_{\mathrm{trunk}}\times\mathcal{M}_{\mathrm{head}}$. 
Over $t=0,\dots,T-1$ communication rounds, the server broadcasts $\theta_t$ to all clients, assumed to participate at every round, and each runs $E$ local epochs on its own data before returning
$\theta_t^{(i)}=(\theta_{\mathrm{trunk},t}^{(i)},\theta_{\mathrm{head},t}^{(i)})$. 
The server then aggregates both blocks. For EEGNet, all parameters are
Euclidean and are averaged via \emph{FedAvg}~\cite{mcmahan2017communication},
\(
  \theta_{t+1}=\frac{1}{N}\sum_{i=1}^{N}\theta_{t}^{(i)}.
\)
For SPDNet, the head is averaged in the same way. The BiMap weights cannot
be, since the arithmetic mean of Stiefel points leaves the manifold.
Following~\cite{pautrel2026fedspdnet},
we use the computationally cheap \emph{ProjAvg} scheme, which maps that mean
back onto the manifold layerwise with the polar factor $\mathrm{uf}(A)=A(A^{\top}A)^{-1/2}$,
\begin{equation}
  W_{\ell,t+1}=\mathrm{uf}\!\left(\frac{1}{N}\sum_{i=1}^{N}W_{\ell,t}^{(i)}\right),
  \qquad \ell=1,\dots,L.
\label{eq:projavg}
\end{equation}
Every learnable parameter is thus transmitted at every round, and all
clients end up with the same model. While this suits clients with similar data distributions, it is less
suited to multi-subject EEG, where spatial patterns, signal quality and discriminative
rhythms vary across subjects.

\textbf{Personalised FL.} We follow FedPer~\cite{arivazhagan2019fedper},
which assumes that clients share a common representation but need their
own decision rule. The trunk is therefore aggregated as in standard FL,
while the head stays local. The model of client $i$ thus combines the
common trunk $\theta_{\mathrm{trunk}}$ with its own head
$\theta_{\mathrm{head}}^{(i)}$. The objective becomes
\[
  \min_{\theta_{\mathrm{trunk}},\,\{\theta_{\mathrm{head}}^{(i)}\}_{i=1}^{N}}
  \frac{1}{N}\sum_{i=1}^{N}
  F_i\!\left(\theta_{\mathrm{trunk}},\theta_{\mathrm{head}}^{(i)}\right).
\]
A round proceeds as in standard FL, except that only the trunk is
exchanged. The server broadcasts $\theta_{\mathrm{trunk},t}$. Client $i$
combines it with its own head $\theta_{\mathrm{head},t}^{(i)}$, trains both
for $E$ local epochs and returns $\theta_{\mathrm{trunk},t}^{(i)}$. The
server aggregates the trunks with \emph{FedAvg} for EEGNet and with
\emph{ProjAvg}~\eqref{eq:projavg} for the BiMap weights of SPDNet. Each head is
trained at every round but never leaves its client.

\section{Experimental design}
\label{sec:design}
\subsection{Compared configurations}
We compare three configurations, listed in Table~\ref{tab:configs}.
\emph{Standard FL} and \emph{personalised FL} are the two federated
protocols of Section~\ref{ssec:fl}. They differ only in the head, which is
aggregated in the first and kept local in the second. Comparing them
isolates the effect of personalising the head on top of a common trunk.
\emph{Centralised} training pools the data of all clients and trains a
single model. It is the usual reference for FL, and we select the
architecture hyperparameters on it (Section~\ref{sec:experiments}).

\begin{table}[t]
  \centering
  \setlength{\tabcolsep}{3pt}
  \renewcommand{\arraystretch}{0.95}
  \caption{Configurations. The last two are federated. \emph{Pooled}: one
    model trained on the data of all clients.}
  \label{tab:configs}
  \begin{tabular}{@{}lc@{\hspace{12pt}}cc@{}}
    \toprule
    & Centralised & Standard FL & Personalised FL \\
    \midrule
    Trunk & pooled & aggregated & aggregated \\
    Head  & pooled & aggregated & local \\
    \bottomrule
  \end{tabular}
\end{table}

\subsection{Setup}
\label{sec:experiments}

\textbf{Datasets.} We use three motor-imagery datasets from the MOABB
benchmark~\cite{jayaram2018moabb} (Table~\ref{tab:setup}). We chose them
to cover varied regimes along three axes. The number of channels sets the
size of the covariance matrices, the number of subjects sets the number of
clients, and the number of classes sets the output size of the head.
All signals are band-pass filtered to $[8,32]$\,Hz and resampled to $128$\,Hz.

\textbf{Models and hyperparameters.} SPDNet uses a single BiRe block of output dimension $d$ throughout and EEGNet uses the architecture of~\cite{lawhern2018eegnet} unchanged. 
Both are trained on the cross-entropy risk with the Adam
optimiser~\cite{kingma2014adam}. For SPDNet, the Stiefel weights are
optimised through tangent-space local
trivialisation~\cite{pmlr-v97-lezcano-casado19a}, with backpropagation
through the eigendecomposition~\cite{ionescu2015matrix}. The values used are collected in Table~\ref{tab:setup}. The BiMap output dimension $d$ and the batch size were selected by grid search on the centralised runs and kept fixed elsewhere. The learning rate is selected separately for centralised and federated training, the latter on standard FL and reused for personalised FL. The rectification threshold $\varepsilon$ is set to the $15$th percentile of the eigenvalues of the input covariance matrices of the first three subjects. It is fixed once
before training and shared by all configurations.

\textbf{Protocol.} Every model is trained on the data of all clients, one client being one subject. 
To limit client drift, each client runs a single local epoch before sending its
parameters to the server. Each subject's trials are per-seed split into class-stratified training,
validation and test sets in $75/10/15$ proportions. Trials are normalised individually, by per-channel standardisation for
EEGNet and to unit trace for the SPDNet covariance matrices, which removes the
recording gain. Every reported number is a mean over ten seeds, each redrawing the split and the initialisation. 

\textbf{Training schedule and evaluation.} 
Training is driven by the validation loss, pooled over clients in federated runs: the
learning rate is halved after $20$ epochs (centralised) or rounds (federated) without
improvement, and training stops after $150$ epochs or rounds without improvement, within a budget of $3000$.
All configurations are reported at the epoch or round of lowest validation loss.
\begin{table}[t]
\centering
\footnotesize
\setlength{\tabcolsep}{3pt}
\renewcommand{\arraystretch}{0.88}
\caption{Datasets and setup. Train trials per subject are given as median
(range). Parameter counts are totals, with the share in the head in
parentheses. Thresholds are in units of $10^{-5}$ and learning rates of
$10^{-3}$.}
\label{tab:setup}
\begin{tabular}{@{}lccc@{}}
\toprule
& BNCI2014-001 & Weibo2014 & Cho2017 \\
\midrule
channels $C$                  & 22 & 60 & 64 \\
subjects $S$                  & 9  & 10 & 52 \\
classes $K$                   & 4  & 7  & 2 \\
train trials / subj.              & 432 (398--460) & 420 (366--444) & 150 (132--194) \\
\midrule
SPDNet param.                 & 900 (61\%)    & 2\,677 (55\%) & 1\,298 (21\%) \\
EEGNet param.                 & 2\,484 (41\%) & 3\,863 (47\%) & 2\,514 (15\%) \\
\midrule
BiRe width $(d,\varepsilon)$ & (16,8.0) & (20,2.0) & (16,6.0) \\
\midrule
\multicolumn{4}{@{}l}{\emph{SPDNet\,/\,EEGNet.}} \\
batch                         & 24\,/\,32 & 16\,/\,16 & 32\,/\,32 \\
centralised lr                & 3\,/\,3    & 1\,/\,1    & 0.3\,/\,10 \\
federated lr                  & 60\,/\,10  & 3\,/\,3   & 1\,/\,10 \\
\bottomrule
\end{tabular}
\end{table}


\begin{table}[t]
  \centering
  \setlength{\tabcolsep}{3pt}
  \renewcommand{\arraystretch}{0.95}
  \caption{Test accuracy (\%, mean$_{\pm\text{std}}$). The last two columns
    are federated. Configurations are defined in Table~\ref{tab:configs}.
    Bold marks the best configuration in each row. Chance levels are 25.0
    (BNCI2014-001), 14.3 (Weibo2014) and 50.0 (Cho2017).}
  \label{tab:results}
  \begin{tabular}{@{}lc@{\hspace{12pt}}cc@{}}
    \toprule
    & Centralised & Standard FL & Personalised FL \\
    \midrule
    \multicolumn{4}{c}{\emph{BNCI2014-001}} \\
    SPDNet & $56.4_{\pm 1.9}$ & $52.6_{\pm 2.1}$ & $\mathbf{68.3}_{\pm 1.6}$ \\
    EEGNet & $\mathbf{63.7}_{\pm 2.0}$ & $45.9_{\pm 3.3}$ & $56.1_{\pm 2.7}$ \\
    \midrule
    \multicolumn{4}{c}{\emph{Weibo2014}} \\
    SPDNet & $49.7_{\pm 0.8}$ & $46.9_{\pm 1.7}$ & $\mathbf{56.8}_{\pm 1.9}$ \\
    EEGNet & $\mathbf{52.0}_{\pm 1.0}$ & $38.4_{\pm 1.2}$ & $41.8_{\pm 2.0}$ \\
    \midrule
    \multicolumn{4}{c}{\emph{Cho2017}} \\
    SPDNet & $63.9_{\pm 0.9}$ & $62.5_{\pm 0.8}$ & $\mathbf{65.9}_{\pm 1.2}$ \\
    EEGNet & $\mathbf{67.9}_{\pm 1.0}$ & $65.6_{\pm 1.0}$ & $64.8_{\pm 2.7}$ \\
    \bottomrule
  \end{tabular}
\end{table}
\input{fig_23}
\section{Analysis}\label{sec:analysis}
Table~\ref{tab:results} reports the test accuracies and
Figure~\ref{fig:curves} the learning curves.

\textbf{Effect of personalisation.}
Standard FL loses accuracy with respect to centralised training, for both
architectures. The drop is moderate for SPDNet, from 1.4 to 3.8 points. It
is large for EEGNet on BNCI2014-001 and Weibo2014, with 17.8 and 13.6
points.

For SPDNet, personalised FL improves on standard FL by 3.4 to 15.7 points
and exceeds centralised training on all three datasets, although the gap
lies within one standard deviation on Cho2017. For EEGNet, it improves on
standard FL by 10.2 and 3.4 points on BNCI2014-001 and Weibo2014, but stays
below the centralised model. On Cho2017, personalisation brings no gain
for EEGNet, and both federated configurations stay below centralised
training.

We believe that these results may be explained, at least in part, by how
the clients are built. Each client is one subject, and EEG signals vary
strongly from one subject to another~\cite{saha2020intra}. A domain shift
between clients is therefore quite possible. In standard FL, as in
centralised training, a single decision rule must fit all subjects. A local
head lets each client adapt its decision rule to its own data, while the
common trunk still learns a representation from all
subjects~\cite{collins2021fedrep}. Personalising the head could thus partly
compensate for the shifts between subjects.

\textbf{Riemannian versus Euclidean decoding.}
In centralised training, EEGNet outperforms SPDNet on all three datasets,
by 2.3 to 7.3 points. Federation reverses this order on BNCI2014-001 and
Weibo2014. SPDNet loses less
than EEGNet when moving from centralised training to standard FL, and gains
more from personalisation. Its personalised FL configuration is the best of
all on BNCI2014-001 and Weibo2014, with 68.3 and 56.8 against 63.7 and 52.0 for
centralised EEGNet. On Cho2017, centralised EEGNet stays ahead, with 67.9
against 65.9. In this setting, SPDNet thus appears more robust to
federation and a better candidate for personalised FL.

\textbf{Convergence and communication.} For SPDNet, personalised FL also
speeds up training. It comes within one point of its reported accuracy
(Table~\ref{tab:results}) in 150 to 270 rounds, against 600 to 810 for
standard FL. For EEGNet, personalised and standard FL both come within one
point in 90 to 200 rounds, so personalisation brings no speed-up. Personalised
FL also reduces communication, since only the trunk is transmitted. For SPDNet,
this cuts the parameters sent at each round by 61\%, 55\% and 21\% on
BNCI2014-001, Weibo2014 and Cho2017 (Table~\ref{tab:setup}). Personalising
the head thus makes SPDNet, already smaller than EEGNet, both more accurate
and cheaper to federate.

\section{Conclusion}\label{sec:conclusion}
On the three motor-imagery datasets we considered, personalising the head
made federated SPDNet more accurate than centralised training, and faster
and cheaper to train than standard FL.  A common
Riemannian representation with subject-specific classifiers thus appears
well suited to inter-subject variability in federated learning. This
design opens two directions for future work. First, a new subject with
unlabelled trials would need its head to be built without labels, for
instance through cross-subject domain adaptation. Second, FL keeps the
trials on their site, but the transmitted parameters may still leak
information about them. Differential privacy bounds this leakage by adding
noise to the transmitted parameters. Since the heads stay on the clients,
only the trunk would need to be privatised, which makes personalised SPDNet
a natural candidate for differentially private FL.
\clearpage 
\section{Compliance with Ethical Standards}
This study was conducted on human subject data made available
in open access through the MOABB benchmark and ethical approval was not required. 
\section{Acknowledgments}
This research was supported by DATAIA Convergence Institute as part of the ``Programme d’Investissement d’Avenir'', (ANR-17-CONV-0003) operated by L2S.

\bibliographystyle{IEEEbib}
\bibliography{refs}

\end{document}